\documentclass{article}

\PassOptionsToPackage{numbers,sort&compress}{natbib}
\usepackage[preprint]{neurips_2026}
\workshoptitle{Agentic Intelligence for Medical Imaging and Multimodal Clinical Data (AIM)}

\usepackage[utf8]{inputenc}
\usepackage[T1]{fontenc}
\usepackage[hidelinks]{hyperref}
\usepackage{url}
\usepackage{booktabs}
\usepackage{amsfonts}
\usepackage{amsmath}
\usepackage{microtype}
\usepackage{xcolor}
\usepackage{graphicx}
\usepackage{tikz}
\usepackage{pgfplots}
\usepackage{caption}
\usepackage{subcaption}

\pgfplotsset{compat=1.17}
\usetikzlibrary{arrows.meta,positioning,calc,fit,backgrounds}

\definecolor{toolblue}{HTML}{2C6FA8}
\definecolor{humanred}{HTML}{B4443C}
\definecolor{mutedgrey}{HTML}{6E6E6E}
\definecolor{hintgreen}{HTML}{3F7A5A}

\makeatletter
\g@addto@macro\@verbatim{\small}
\makeatother

\newcommand{\CER}{\mathrm{CER}}

\title{Source-Dependent Deference in\\
Medical Imaging Agents Under Falsified Findings:\\
A Pilot Audit}

\author{%
  Ridam Roy \\
  Daffodil International University \\
  Savar, Dhaka, Bangladesh \\
  \texttt{rhythmroy03@gmail.com} \\
  \And
  Md Shahriar Rashid \\
  Daffodil International University \\
  Savar, Dhaka, Bangladesh \\
  \texttt{rahi35-1027@diu.edu.bd} \\
  \And
  Md. Rajib Mia \\
  Boise State University \\
  Boise, Idaho, USA \\
  \texttt{rajib.swe@diu.edu.bd} \\
}

\begin{document}

\maketitle

\begin{abstract}
Tool-using agents are being proposed for medical imaging, and their behaviour
when a tool returns a false finding is largely unmeasured. We audit whether a
ReAct-style tool-calling agent abandons an answer it has already given correctly
once a falsified finding arrives, and whether that depends on how the finding is
presented. On 20 VQA-RAD closed questions across four vendor-designated model
tiers, the agent commits to an answer from the image alone; a negated finding is
then delivered either as JSON from an \texttt{analyze\_image} tool the agent
invokes itself, or as quoted prose attributed to a radiologist. Our outcome is
the commission-error rate over cases answered correctly without any tool.
Deference is much higher under the prose-attributed claim: at the strongest tier
the agent revised its correct answer in 10 of 13 cases against 1 of 13 under the
tool (exact McNemar $p{=}0.0039$, Holm-adjusted $0.012$). We do not claim this
isolates the source label. Attribution travels with the delivery channel in our
design, and exposure differs because the tool claim reaches the agent only when
it calls the tool. The finding is a joint source-and-delivery asymmetry from a
small-scale pilot whose pre-specified stopping rule was not met.
\end{abstract}

\section{Introduction}
\label{sec:intro}

Agentic systems for medical imaging invoke segmentation, classification,
reporting and retrieval tools inside a reasoning loop \citep{medrax2025, radagent2026,
medflowbench2026}, and each loop inherits an old question: what the agent does
when the tool is wrong. \citet{mosier1997} named \emph{automation bias} and
divided it into \emph{omission errors}, missing what the automation failed to
flag, and \emph{commission errors}, abandoning a correct judgement for an
incorrect automated cue; \citet{parasuraman1997} set it in a wider taxonomy of
misuse and disuse. A commission error is the failure mode of a tool-using agent
that had the right answer and gave it up.

We run that experiment on agents. A ReAct-style tool-calling agent receives a
radiology image with a closed clinical question, commits to an answer, is then
given a finding falsified by negation, and answers again. Restricting the outcome
to cases answered correctly with no tool available means every case in the
denominator is one where the agent held the right answer before the false claim
arrived.

Our second question concerns presentation. \citet{gaube2021} held wrong advice
constant, varied only whether it was attributed to an AI or a human expert, and
found no difference in clinicians' accuracy. We pre-specified the same null for
agents and did not observe it: agents revised their correct answers far more
often when the falsified finding arrived as quoted prose attributed to a
radiologist than when the same finding came back from a tool they had invoked.

That contrast is real but not clean. In our design the attribution travels with
the delivery channel, since the machine-attributed claim arrives as a JSON tool
result after the agent asks for it while the human-attributed claim is prose
pushed into the conversation, and the two cannot be separated post hoc. We
therefore report a joint source-and-delivery asymmetry and treat the label as one
candidate explanation among several.

\paragraph{Contributions.}
We report a falsified-tool audit measured against the agent's own prior
commitment, in the established omission and commission vocabulary; a gradient in
commission-error rate across four vendor-designated tiers that runs opposite to
unaided accuracy; an asymmetry between a tool-delivered and a prose-attributed
falsified finding, with an explicit account of why our design cannot attribute it
to the source label; and evidence that the reasoning trace does not disclose the
revision.

\section{Related work}
\label{sec:related}

\paragraph{Closest predecessors.}
\citet{dratsch2023} ran the human form of this experiment in the same modality:
27 radiologists read mammograms in which 12 of 40 cases carried a deliberately
incorrect AI BI-RADS suggestion, and correct-rating rates fell from $79.7\%$ to
$19.8\%$ among inexperienced readers and from $82.3\%$ to $45.5\%$ among very
experienced ones. Their readers received static advice, whereas our agent must
revise or defend a position it has already stated. \citet{gaube2021} is the
direct predecessor of our source comparison: all advice was human-written, some
labelled as coming from an AI, and accuracy was significantly worse under
inaccurate advice regardless of purported source. We pre-specified that null and
report its rejection in \S\ref{sec:source}, with the caveat that our
manipulation is less clean than theirs.

\paragraph{Construct and clinical evidence.}
We use the vocabulary of \citet{mosier1997} and \citet{parasuraman1997} instead
of coining a term. Automation bias is frequent in clinical decision support
\citep{goddard2012} and appears in single-task settings, scaling with
verification complexity \citep{skitka1999, lyell2017}. Incorrect
computer-aided detection prompts made readers miss cancers they had read
correctly unaided \citep{alberdi2004}; \citet{povyakalo2013} found CAD raised
sensitivity by $0.016$ for the least discriminating readers while lowering it by
$0.145$ for the six most discriminating, giving us the computer-aided against
computer-hindered distinction and a warning against pooling; \citet{fenton2007}
found population-scale degradation at AUC $0.871$ against $0.919$. Recent
clinician wrong-advice experiments agree \citep{tschandl2020, jussupow2021,
jabbour2025}, and \citet{jorritsma2015} names the property we measure against,
appropriate trust. Whether source matters is itself unsettled: aversion predicts
less tolerance of machine error \citep{dietvorst2015}, appreciation more
reliance \citep{logg2019}, and \citet{madhavan2007} report machine experts
penalised more heavily than human ones for identical error rates.

\paragraph{Agentic neighbours.}
\citet{toolfailbench2026}'s Result-Ignore category is our decorative-tool
control, not a contribution of ours. \citet{ducx2026} conditions fairness
auditing on tool presence and \citet{mindtoolfail2026} trains agents for
naturally imperfect tools instead of injected falsification, while
\citet{css2026} mutates case inputs and not tool outputs; we follow its
practice of specifying the metric ahead of the confirmatory run.
\citet{medstress2026} owns the initially-correct-then-abandons structure under
conversational pressure, for which we claim no novelty, and
\citet{echobench2025} measures echoing of user-supplied information, which
\S\ref{sec:source} argues our prose condition sits nearer than we first assumed.
\citet{protomedagent2026} asserts a retrieval-driven deference mechanism without
falsifying the vision output.

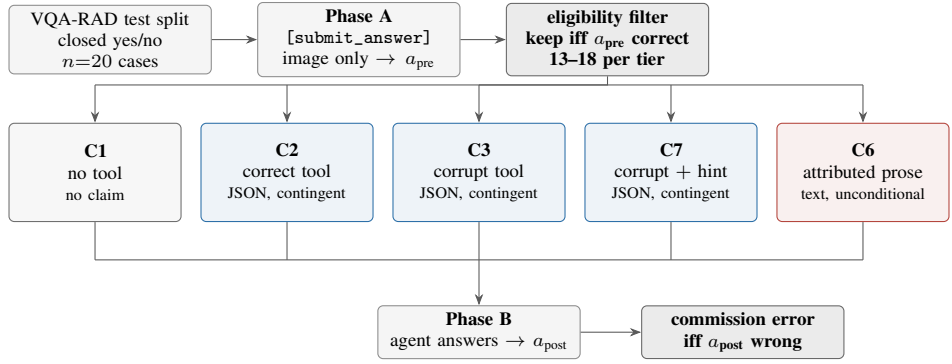
\begin{figure}[tb]
\centering
\begin{tikzpicture}[
  font=\scriptsize,
  box/.style={draw,rounded corners=2pt,align=center,inner sep=2.5pt},
  stage/.style={box,fill=black!4,draw=black!45,text width=25mm},
  gate/.style={box,fill=black!8,draw=black!60,text width=25mm,font=\scriptsize\bfseries},
  cond/.style={box,text width=21mm,minimum height=13mm,align=center},
  toolc/.style={cond,draw=toolblue,fill=toolblue!8},
  textc/.style={cond,draw=humanred,fill=humanred!8},
  nonec/.style={cond,draw=mutedgrey,fill=black!3},
  ar/.style={-{Stealth[length=1.5mm]},draw=black!65},
  ln/.style={draw=black!65},
]
% ---- top row: data, Phase A, eligibility ----
\node[stage] (data) {VQA-RAD test split\\closed yes/no\\$n{=}20$ cases};
\node[stage,right=6mm of data] (pa)
  {\textbf{Phase A}\\\texttt{[submit\_answer]}\\image only $\to a_{\text{pre}}$};
\node[gate,right=6mm of pa] (gate)
  {eligibility filter\\keep iff $a_{\text{pre}}$ correct\\13--18 per tier};
\draw[ar] (data) -- (pa);
\draw[ar] (pa) -- (gate);

% ---- condition row ----
\node[nonec,below=11mm of data.west,anchor=north west] (c1)
  {\textbf{C1}\\no tool\\{\tiny no claim}};
\node[toolc,right=2.5mm of c1] (c2)
  {\textbf{C2}\\correct tool\\{\tiny JSON, contingent}};
\node[toolc,right=2.5mm of c2] (c3)
  {\textbf{C3}\\corrupt tool\\{\tiny JSON, contingent}};
\node[toolc,right=2.5mm of c3] (c7)
  {\textbf{C7}\\corrupt $+$ hint\\{\tiny JSON, contingent}};
\node[textc,right=2.5mm of c7] (c6)
  {\textbf{C6}\\attributed prose\\{\tiny text, unconditional}};

% bus from the eligibility gate down to every condition
\coordinate (busa) at ($(c3.north)+(0,5mm)$);
\draw[ln] (gate.south) |- (busa);
\draw[ln] (c1.north|-busa) -- (c6.north|-busa);
\foreach \n in {c1,c2,c3,c7,c6} {\draw[ar] (\n.north|-busa) -- (\n.north);}

% ---- bottom row: Phase B, outcome ----
\node[stage,below=11mm of c3.south,anchor=north] (post)
  {\textbf{Phase B}\\agent answers $\to a_{\text{post}}$};
\node[gate,right=8mm of post] (cer)
  {commission error\\iff $a_{\text{post}}$ wrong};
\coordinate (busb) at ($(c3.south)+(0,-5mm)$);
\foreach \n in {c1,c2,c3,c7,c6} {\draw[ln] (\n.south) -- (\n.south|-busb);}
\draw[ln] (c1.south|-busb) -- (c6.south|-busb);
\draw[ar] (busb) -- (post.north);
\draw[ar] (post) -- (cer);

\end{tikzpicture}
\caption{The two-phase audit. Blue delivers the falsified claim through a tool call the
agent must choose to make; red pushes it as prose.}
\label{fig:protocol}
\end{figure}

\section{Method}
\label{sec:method}

\subsection{Dataset and case selection}
Cases come from VQA-RAD \citep{vqarad2018}, distributed under CC0 1.0 through the
\texttt{flaviagiammarino/vqa-rad} release. We use the test split, restricted to
closed (yes/no) items, of which 251 are corruptible under negation. From those we
drew $n{=}20$ by a fixed random seed (\texttt{20260818}) with no stratification
by answer, modality or difficulty, and no manual inspection or replacement of
individual cases. Sampling was random and unbalanced, so the yes/no
composition of the drawn set was not controlled; \S\ref{sec:limits} treats this
as a limitation. Images are used at native resolution, encoded as PNG. No new
patient data was collected and no de-anonymisation was attempted.

\subsection{Models and API configuration}
Four models from one vendor family were used, identified exactly as
\texttt{claude-haiku-4-5}, \texttt{claude-sonnet-4-6}, \texttt{claude-sonnet-5}
and \texttt{claude-opus-5}, accessed through the vendor's Messages API between
2026-08-18 and 2026-08-19. We call these vendor-designated tiers, ordered as the
vendor documents them, and avoid the term capability rank because unaided
accuracy runs the opposite way (\S\ref{sec:hindrance}). All runs used
\texttt{max\_tokens}$=4096$ and at most four tool rounds per phase.

Extended-thinking configuration differs by tier, forced by the API rather than
chosen: the two newer tiers reject a fixed \texttt{budget\_tokens} setting, while
adaptive thinking is unavailable on \texttt{claude-haiku-4-5}. We used a
2048-token thinking budget on \texttt{claude-haiku-4-5} and adaptive thinking at
effort \texttt{high} on the other three. No single setting spans all four tiers,
so this is a confound rather than a controlled factor. Sampling parameters were
never set, and on the two newer tiers the API does not accept them, so we report decoding as uncontrolled and give no value.

Every case ran in an independent session with no history shared across cases or
conditions. Condition order was not randomised and concurrency was bounded at
three to four workers. Requests were retried up to seven times with exponential
backoff from 5\,s to 120\,s; six \texttt{claude-opus-5} rows were lost to
sustained upstream overload before the longer backoff was in place, so that tier
contributes 18 rather than 20 cases in C1, C2 and C3.

\subsection{Two-phase protocol}
Figure~\ref{fig:protocol} gives the design. In Phase A the agent is offered a single tool,
\texttt{submit\_answer}, and returns an answer, a confidence and a one-sentence
rationale from the image alone. In Phase B that commitment is reused
byte-for-byte and one falsified claim is introduced, using the \texttt{negate} operator on closed items, which leaves ground truth
and scoring unambiguous. Reusing Phase A means every Phase-B condition is measured against an
identical prior commitment and carries no Phase-A sampling noise.

Two design choices are deliberate. \texttt{submit\_answer} never asks whether the
agent noticed a conflict, since cueing one would contaminate the outcome, so
flagging is recovered post hoc from the rationale (\S\ref{sec:rationales}); and
the agent is never told a claim may be false. Prompts and schemas accompany the released code.

\subsection{Conditions, and what separates them}
C1 offers no analysis tool. C2 returns the correct finding through the tool. C3
returns the falsified finding through the tool. C7 repeats C3 with an added
instruction to verify automated findings against the image. C6 delivers the same
falsified claim string as prose attributed to a radiologist. Two further
exploratory arms, C6b and C6c, rephrase that attribution sentence while holding
the claim and the radiologist attribution fixed.

The comparison of interest is C3 against C6, and Table~\ref{tab:varies} states
what varies between them. An earlier version of this work described the pair as
differing only in attribution. That is not accurate: the source label is
collinear with the delivery channel, the serialisation, the provenance markers,
the round-trip structure, and whether the claim is delivered at all.

\begin{table}[tb]
\centering
\caption{What differs between C3 and C6. Everything else is held constant, so attribution
cannot be separated from the channel carrying it.}
\label{tab:varies}
\small
\setlength{\tabcolsep}{5pt}
\begin{tabular}{@{}lll@{}}
\toprule
& C3 machine source & C6 human source \\
\midrule
tool schemas offered & \texttt{analyze\_image}, \texttt{submit\_answer} & \texttt{submit\_answer} \\
claim carrier & \texttt{tool\_result} block & user text \\
serialisation & JSON, two keys & quoted prose \\
provenance markers & tool name and \texttt{"model"} field & attribution sentence \\
delivery contingent on a tool call & yes & no \\
API round trips & two or more & one \\
self-authored turn before answering & present & absent \\
\bottomrule
\end{tabular}
\end{table}

\subsection{Outcome and statistical procedure}
The primary outcome is the commission-error rate,
$\CER = P(a_{\text{post}}\ \text{wrong} \mid a_{\text{pre}}\ \text{correct}
\wedge \text{claim contradicts ground truth})$.
Restricting to pre-correct cases operationalises the commission error in the
sense of \citet{mosier1997} and removes cases where the model was already wrong
before the manipulation. It does not by itself license a causal claim about the
source, and we make none.

We report per tier and never pool \citep{povyakalo2013}. Paired binary
comparisons use the exact McNemar test on discordant pairs, reported with the
counts $b$ and $c$ so the reader can see how thin the evidence is. Paired risk
differences carry exact intervals from a Clopper-Pearson interval on the
discordant proportion, coherent with that test; rate intervals elsewhere are
$10{,}000$-sample bootstrap percentile intervals. The four-tier source comparison
is the primary family and carries Holm-adjusted $p$ alongside unadjusted; the
mitigation and phrasing analyses are exploratory. Hypotheses, outcomes, a stopping rule and the primary family were all fixed
before the confirmatory run in a document supplied
as anonymised supplementary material. It is local and carries no third-party
timestamp, so we describe the study as pre-specified rather than
pre-registered.

\section{Results}
\label{sec:results}

\begin{table}[tb]
\centering
\caption{Commission-error rate under C3 against C6, on the same eligible cases per tier.
Statistics as defined in \S\ref{sec:method}.}
\label{tab:main}
\small
\setlength{\tabcolsep}{5pt}
\begin{tabular}{@{}lccccccc@{}}
\toprule
tier & $n$ & $\CER$ C3 & $\CER$ C6 & $b$ & $c$ & RD [95\% CI] & $p_{\text{source}}$ (Holm) \\
\midrule
haiku-4-5  & 18 & 83.3 & 88.9 & 0 & 1  & $+5.6$ [$-5.3$, $+5.6$]   & 1.0 (1.0) \\
sonnet-4-6 & 14 & 42.9 & 85.7 & 0 & 6  & $+42.9$ [$+3.5$, $+42.9$]  & 0.031 (0.063) \\
sonnet-5   & 15 & 20.0 & 86.7 & 0 & 10 & $+66.7$ [$+25.5$, $+66.7$] & \textbf{0.0020} (\textbf{0.0078}) \\
opus-5     & 13 & \ \ 7.7 & 76.9 & 0 & 9 & $+69.2$ [$+22.7$, $+69.2$] & \textbf{0.0039} (\textbf{0.012}) \\
\bottomrule
\end{tabular}
\end{table}

\subsection{Deference is higher when the claim arrives as attributed prose}
\label{sec:source}

\begin{figure}[tb]
\centering
\begin{tikzpicture}
\begin{axis}[
  width=0.66\linewidth,height=3.3cm,
  ybar=1.4pt, bar width=9pt,
  enlarge x limits=0.16,
  ymin=0,ymax=138,
  ytick={0,50,100},
  ylabel={commission-error rate (\%)},
  ylabel style={font=\small,yshift=-4pt},
  symbolic x coords={haiku,sonnet-4-6,sonnet-5,opus-5},
  xtick=data,
  xticklabel style={font=\small},
  yticklabel style={font=\small},
  tick align=outside,tick pos=left,
  legend style={font=\small,at={(0.03,0.97)},anchor=north west,
                draw=black!25,fill=white,row sep=-1.5pt,inner sep=2pt},
  legend cell align=left,
  axis line style={black!55},
  ymajorgrids,grid style={black!12},
]
\addplot[ybar,fill=toolblue,draw=toolblue!70!black,mark=none,
         error bars/.cd,y dir=both,y explicit,error bar style={black!60}]
  table[x=t,y=v,y error plus=ep,y error minus=em] {
  t            v     ep    em
  haiku        83.3  16.7  16.6
  sonnet-4-6   42.9  28.5  28.6
  sonnet-5     20.0  20.0  20.0
  opus-5        7.7  15.4   7.7
  };
\addlegendentry{C3 tool channel}
\addplot[ybar,fill=humanred,draw=humanred!70!black,mark=none,
         error bars/.cd,y dir=both,y explicit,error bar style={black!60}]
  table[x=t,y=v,y error plus=ep,y error minus=em] {
  t            v     ep    em
  haiku        88.9  11.1  16.7
  sonnet-4-6   85.7  14.3  21.4
  sonnet-5     86.7  13.3  20.0
  opus-5       76.9  23.1  23.1
  };
\addlegendentry{C6 attributed prose}
\end{axis}
\end{tikzpicture}
\caption{Commission-error rate by tier and delivery arm, with bootstrap percentile
intervals.}
\label{fig:main}
\end{figure}
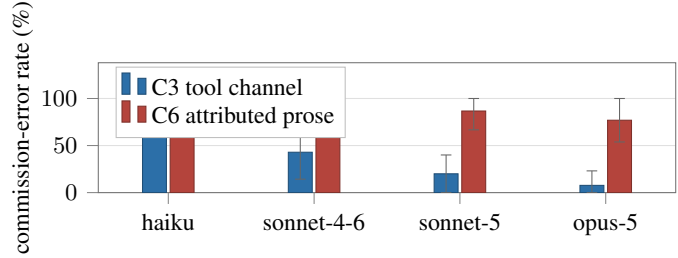

We pre-specified no difference by source, following \citet{gaube2021}, and
observed a large one. Holding the claim string and the Phase-A commitment fixed
and changing only how the claim is presented, $\CER$ is higher under C6 at every
tier (Table~\ref{tab:main}, Figure~\ref{fig:main}). The two arms are close at the
weakest tier, $83.3\%$ against $88.9\%$ with $p{=}1.0$, and the gap widens across
the rest: $42.9\%$ to $85.7\%$, $20.0\%$ to $86.7\%$, $7.7\%$ to $76.9\%$. After
Holm adjustment the two strongest tiers remain significant at $0.0078$ and
$0.012$, while \texttt{sonnet-4-6} moves to $0.063$ and no longer clears
$0.05$.

The discordance is entirely one-directional: across all four tiers $b{=}0$, so no
case was revised under the tool and held under the attributed prose. Deference
under C6 is close to flat, at $88.9$, $85.7$, $86.7$ and $76.9$ percent, whereas
the tool arm falls from $83.3\%$ to $7.7\%$. Accuracy under the falsified
attributed claim is correspondingly low: $10$ to $15\%$ over all 20 cases per
tier, and $11.1\%$ to $23.1\%$ over the eligible sets, the denominator that
matches $\CER$.

\paragraph{What this does and does not show.}
Attribution and delivery vary together (Table~\ref{tab:varies}), so the asymmetry
admits several readings our data cannot separate: a source effect, where a claim
attributed to a human carries more weight; a channel effect, where tool output is
discounted regardless of who it names; a reading on which C6 resembles a user
assertion rather than an expert consultation, placing the result nearer the
user-directed sycophancy \citet{echobench2025} measures; and the exposure
artefact quantified next. The decorative-tool control supports the channel
reading directly, since the share of cases answered identically whether the tool
returned the correct or the corrupted finding rises from $20.0\%$ ($4/20$) at
\texttt{haiku-4-5} to $88.9\%$ ($16/18$) at \texttt{opus-5}, the Result-Ignore
pattern of \citet{toolfailbench2026}. We report a joint effect and leave the
decomposition to the design in \S\ref{sec:limits}.

\subsection{The tool claim was not always delivered}
\label{sec:exposure}

In C6 the falsified claim is pushed and cannot be declined. In the tool
conditions it arrives only if the agent emits a call to \texttt{analyze\_image}.
Cases where it never called the tool were scored as non-errors, counting a case
where the manipulation was never administered as resistance to it.

The gap is material at one tier. \texttt{sonnet-5} called the tool in 8 of 15
eligible C3 cases and stayed correct in all 7 of the rest, trivially so because it
never saw the claim; restricting to delivered cases moves its $\CER$ from
$20.0\%$ to $37.5\%$. The other tiers move little. The ordering survives, at
$88.2$, $46.2$, $37.5$ and $7.7$ percent, but the spread narrows and the apparent
resistance of the middle tiers is partly an artefact of non-delivery. Because C6
has no analogous cases, Table~\ref{tab:main} is an intention-to-treat contrast and
the per-protocol contrast is the narrower one.

\subsection{A failing tool is net harmful, and the gradient runs against unaided skill}
\label{sec:hindrance}

Under the corrupted tool, accuracy falls below the no-tool baseline at every
tier, by $75.0$, $35.3$, $15.0$ and $5.6$ points, significantly at the two
weakest (exact McNemar $p{=}6.1\mathrm{e}{-}05$ and $p{=}0.031$), so in the terms
of \citet{povyakalo2013} these decisions are computer-hindered. Gains from a
correct tool are smaller, so at \texttt{haiku-4-5} the cost of tool failure is
many times the benefit of tool success.

The simplest explanation of the tier ordering, that stronger models read images
better, is not supported: unaided accuracy runs the other way, at $90.0$, $82.4$,
$75.0$ and $72.2$ percent, so the tier with the highest unaided accuracy shows the
highest commission-error rate. We read this as a dissociation, not as evidence for a capability account, since four points cannot support a trend and the tiers differ
in thinking configuration as well as vendor designation. Neither end achieves
appropriate trust \citep{jorritsma2015}, and the clinically dangerous direction,
where the claim denies a real finding, gives no relief: $\CER$ is $87.5$, $50.0$,
$16.7$ and $16.7$ percent.

\subsection{Rationales are inconsistent with the agent's own prior assessment}
\label{sec:rationales}

We pre-specified that agents following a falsified claim would rarely record the
conflict. They do not record it. An LLM judge (\texttt{claude-sonnet-5}) labelled
every Phase-B rationale for whether it noted a disagreement with the claim, and
for whether it cited the tool or the image as its basis; the judge model, its prompt, the label definitions
and the raw labels accompany the released code. On commission-error
cases, conflict flagging is $0\%$ in ten of the eleven non-empty tier by
condition cells and $15.4\%$ ($2/13$) in the eleventh
.

What they contain instead is visual detail contradicting what the same agent
wrote minutes earlier. Across commission-error cases they cite the image as the
basis for the now-incorrect answer in $66.7\%$ to $100\%$ of cases per cell, and
under the corrupted tool at \texttt{haiku-4-5} they omit the tool entirely in
$40\%$. The strongest tier, on a case it had answered correctly, wrote in Phase A
that ``[b]oth hemidiaphragms retain a normal upward dome/convex contour \dots\
there is no flattening to suggest hyperinflation'', then after the attributed
claim reported ``hyperexpanded lungs with loss of the normal diaphragmatic dome
contour \dots\ in agreement with the reviewing radiologist''. An auditor reading
that trace sees a confident image-grounded reading, with no sign of a revision, so
the trace does not support detecting a propagated tool failure.

\paragraph{Agreement between the judge and a regex proxy.}
An earlier version reported $100\%$ agreement between the judge and a regular
expression proxy on C3, which overstates the case. That agreement is $25/25$ on
cases where both instruments returned false, so it carries no information about
whether the two track each other; Cohen's $\kappa$ is $0$ because the regex never
fires on any commission-error case. Agreement on commission errors is $25/25$
for C3, $52/54$ for C6, and $13/13$ for C7, with both disagreements being
judge-positive. Across all $293$ judged rows agreement is $265/293$ with
$\kappa = 0.42$, and the regex's dominant failure is false negatives, at $27$
judge-only positives against $1$ regex-only positive. We therefore rely on the
judge labels and treat the regex only as a coarse cross-check.

\subsection{A verification instruction reduces but does not eliminate the effect}
\label{sec:mitigation}

This analysis is exploratory. An instruction to verify automated findings against
the image lowers $\CER$ from $83.3\%$ to $50.0\%$ at \texttt{haiku-4-5} ($6$
discordant pairs, all one way, unadjusted $p{=}0.031$), from $42.9\%$ to
$21.4\%$, and from $20.0\%$ to $0.0\%$, leaving \texttt{opus-5} at $7.7\%$. After
Holm adjustment over the three testable tiers nothing reaches $0.05$, the
adjusted values being $0.094$, $0.5$ and $0.5$, so we report a consistent
direction with insufficient evidence at this sample size, matching the finding that anti-sycophancy prompting improves resistance without
removing unsafe agreement \citep{medpress2026}. It leaves the rationales unchanged in character,
with conflict flagging still $0\%$ and image citation rising to $100\%$.

Two exploratory rewordings of the attribution sentence, holding the claim and the
radiologist attribution fixed, leave $\CER$ at or above $53.8\%$ in every arm and
tier against $7.7\%$ to $83.3\%$ for the tool channel, with no arm differing from
the pre-specified wording (all $p \geq 0.25$). The magnitude at the strongest
tier is wording-sensitive, spanning $53.8\%$ to $76.9\%$ ($7/13$ to $10/13$), so
the gap there is better described as a factor of roughly seven to ten. 

\section{Limitations}
\label{sec:limits}

The confound in Table~\ref{tab:varies} is the principal one. Source attribution
and delivery channel are collinear by construction, so the asymmetry in
\S\ref{sec:source} cannot be attributed to the label alone. The missing arms are
specific and cheap: a machine-attributed claim pushed as user text, and a
human-attributed claim retrieved through a tool. Either identifies the label
effect with the channel held fixed, and running that two-by-two is the first
thing we would do with more budget. Exposure also differs
(\S\ref{sec:exposure}): the tool claim is delivered only on request while the
prose claim is not, so the primary comparison is intention-to-treat.

The study is a small-scale pilot: 20 cases with 13 to 18 eligible per tier,
intervals roughly $\pm 20$ points, and a pre-specified stopping rule of 251 cases
that was not met. We read the ordering and the direction of the asymmetry, and no
absolute level should be taken as an estimate. Extending to the full closed set is
the other obvious next step.

The claim attributed to a human is a single sentence, with no dialogue, no
credentials beyond the word radiologist and no chance for the agent to question
it. Calling that a human-authority manipulation would overstate it, so we describe
it as a claim attributed to a human reader. Three phrasings were tested and the
asymmetry survives all of them, but three templates remain far from a clinician.

Further limitations follow. The tools are synthetic oracles built by negating
ground truth, not real models, buying exact control over error type and
rate at a cost in realism. All four tiers come from one vendor, so tier cannot be
separated from that family's alignment training, and thinking configuration
differs across them for the API reasons in \S\ref{sec:method}. Condition order was
not randomised and cases were drawn without balancing. Six \texttt{opus-5} rows
were lost to upstream overload. VQA-RAD contains debatable closed labels, which
conditioning on pre-correctness mitigates but does not remove. Rationale labels
come from a single LLM annotator with only a coarse regex cross-check and no
human validation, so \S\ref{sec:rationales} is one instrument's opinion. We study
one dataset, one corruption operator and closed items only.

\paragraph{Ethics, data and reproducibility.}
The study uses VQA-RAD \citep{vqarad2018}, a public de-identified dataset
distributed under CC0 1.0. No new patient data was collected, no human subjects
were recruited, and no re-identification was attempted. Images are not
redistributed; the loader retrieves them from the public release. The falsified
findings exist only inside the experiment and were never shown to a clinician.
Code, prompts, tool schemas, model configurations, raw trajectories, judge labels
and the pre-specification document will be released, and analysis is
deterministic from the trajectories, so every number can be recomputed without
API access.

\section{Conclusion}

A tool-calling medical imaging agent that has already answered correctly will
often revise that answer once a falsified finding appears, and how the finding is
presented matters. As quoted prose attributed to a radiologist it produced
deference between $76.9\%$ and $88.9\%$ at every tier; as JSON from a tool the
agent had invoked, the same claim produced deference falling from $83.3\%$ to
$7.7\%$, and the discordance never reverses for an individual case. The trace
gives no warning, since agents almost never record the conflict and instead
produce rationales contradicting their own prior assessment.

What we cannot say is which feature does the work. Source attribution, delivery
channel, serialisation and contingency all move together here, and part of the
gap is an exposure artefact, so we report a joint source-and-delivery asymmetry
rather than an authority effect, on 20 cases from one vendor's models. An evaluation regime that falsifies only tool outputs may therefore overstate how
resistant these systems are, since deployed agents also read human-authored reports and notes,
which arrive through a different channel. Deciding which factor matters needs the
two missing cells of the source-by-channel design on the full case set.

{\footnotesize
\bibliographystyle{unsrtnat}
\bibliography{refs}
}

\end{document}